\documentclass{article}
\usepackage{spconf,amsmath,graphicx,newtxmath}

\title{Learning Expressive and Generalizable Motion Features for Face Forgery Detection}
\name{Jingyi Zhang\sthanks{Equal contribution.}, Peng Zhang\footnotemark[1], Jingjing Wang, Di Xie\sthanks{Corresponding author (xiedi@hikvision.com).}, Shiliang Pu}
\address{
Hikvision Research Institute
}

\begin{document}
%
\maketitle
\begin{abstract}
Previous face forgery detection methods mainly focus on appearance features, which may be easily attacked by sophisticated manipulation. Considering the majority of current face manipulation methods generate fake faces based on a single frame, which do not take frame consistency and coordination into consideration, artifacts on frame sequences are more effective for face forgery detection.
However, current sequence-based face forgery detection methods use general video classification networks directly, which discard the special and discriminative motion information for face manipulation detection.
To this end, we propose an effective sequence-based forgery detection framework based on an existing video classification method.
To make the motion features more expressive for manipulation detection, we propose an alternative motion consistency block instead of the original motion features module.
To make the learned features more generalizable, we propose an auxiliary anomaly detection block.
With these two specially designed improvements, we make a general video classification network achieve promising results on three popular face forgery datasets.
\end{abstract}
\section{Introduction}
\label{sec:intro}
In recent years, an increasing number of face forgery videos generated by Deepfake methods caused serious social problems.
Therefore, it is essential to propose an efficient framework to distinguish the real face from forgery productions.
\begin{figure}[t]
\centering
\includegraphics[width=0.9\columnwidth]{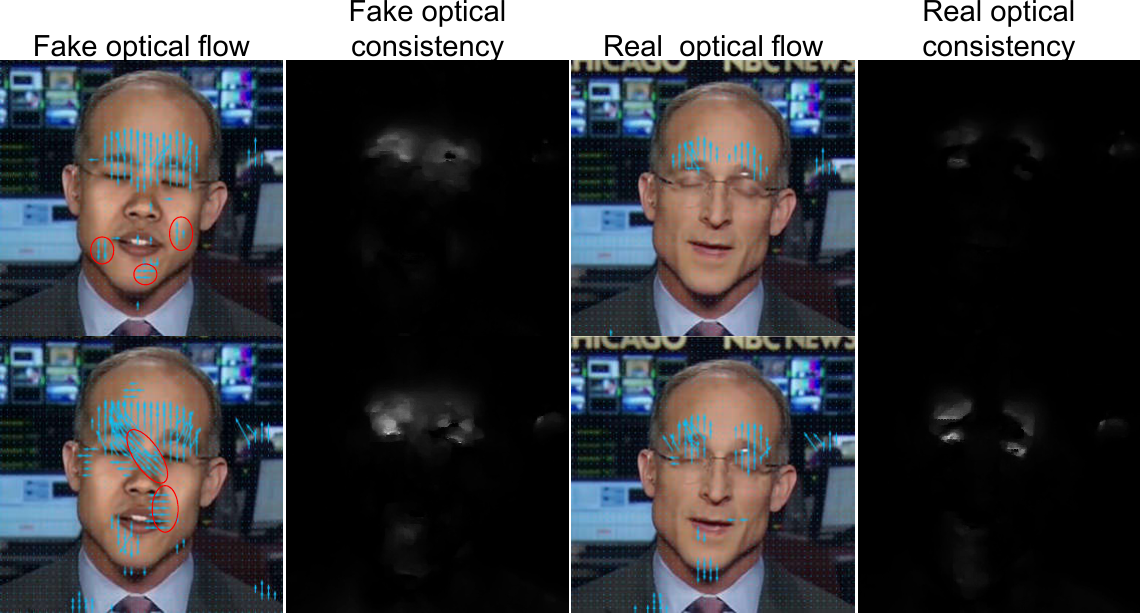}
\caption{Visualization of motion information in real and fake faces.
We visualize the intensity and direction of optical flow for fake faces(col1) and real faces(col3) every 20 pixels.
}
\label{fig1}
\vspace{-0.5cm}
\end{figure}
So far, face forgery detection methods could be roughly divided into two types, namely image-based methods and sequence-based methods. The image-based methods distinguish fake faces by individual frames.
Generally, they leverage a well designed backbone to extract features of a face image.
Classification or clustering losses are deployed to guide these backbones to focus on textual artificial traces e.g. image blending clues
and unusual facial decoration.
\cite{FFPP,DFO,CDF,sun2021domain,Xception} implement several popular CNN backbones to extract image-level features and a binary classifier is applied to distinguish the manipulated faces from the real ones.
Despite the expressive backbones, \cite{Matt,qian2020thinking,chen2021local,zhang2019detecting} design some texture and frequency aware architectures for better forgery clue detection.
Existing frame-based face forgery detection frameworks mainly pay attention to appearance and high-frequency features and still suffer from sophisticated manipulation which leaves fewer appearance anomaly clues.

The sequence-based methods mainly use general video classification networks directly.
\cite{2018Deepfake} uses a universal 2D CNN backbone to perform spatial feature extraction for a frame sequence, and an RNN to model the temporal relationship between the spatial features.

\cite{2020Deepfake,DFO,2020Spatio,WildDeepfake} utilize a 3D convolutional backbone to extract spatio-temporal features directly for a frame sequence.
Current sequence-based methods do not show obvious advantages against frame-based methods.
We think the main reasons are two folds. Firstly, current sequence-based methods are designed originally for general video classification which are lack of effective modeling for face forgery detection.
{
Secondly, current forgery detection methods pay overzealous attention to regularizing the feature learning in the spatial domain,
while what is the effective regularization mechanism in the motion domain is still underexplored.
}

Existing forgery methods generate forge faces by individual frames and there exists little constraint between frames. The lack of consistency constraints between frames leads to unnatural expressions and facial organ movements. As presented in the first column of Fig. \ref{fig1}, there exist obvious unexpected optical flows in manipulated regions(eyes, mouth). The second column and fourth column of Fig. \ref{fig1} show that motion inconsistency in fake faces is much stronger than in real faces. Based on this observation, we propose a motion consistency block to leverage the characteristics of face forgery methods.

\begin{figure*}[t]
\centering
\includegraphics[width=1\textwidth,height=0.33\textheight]{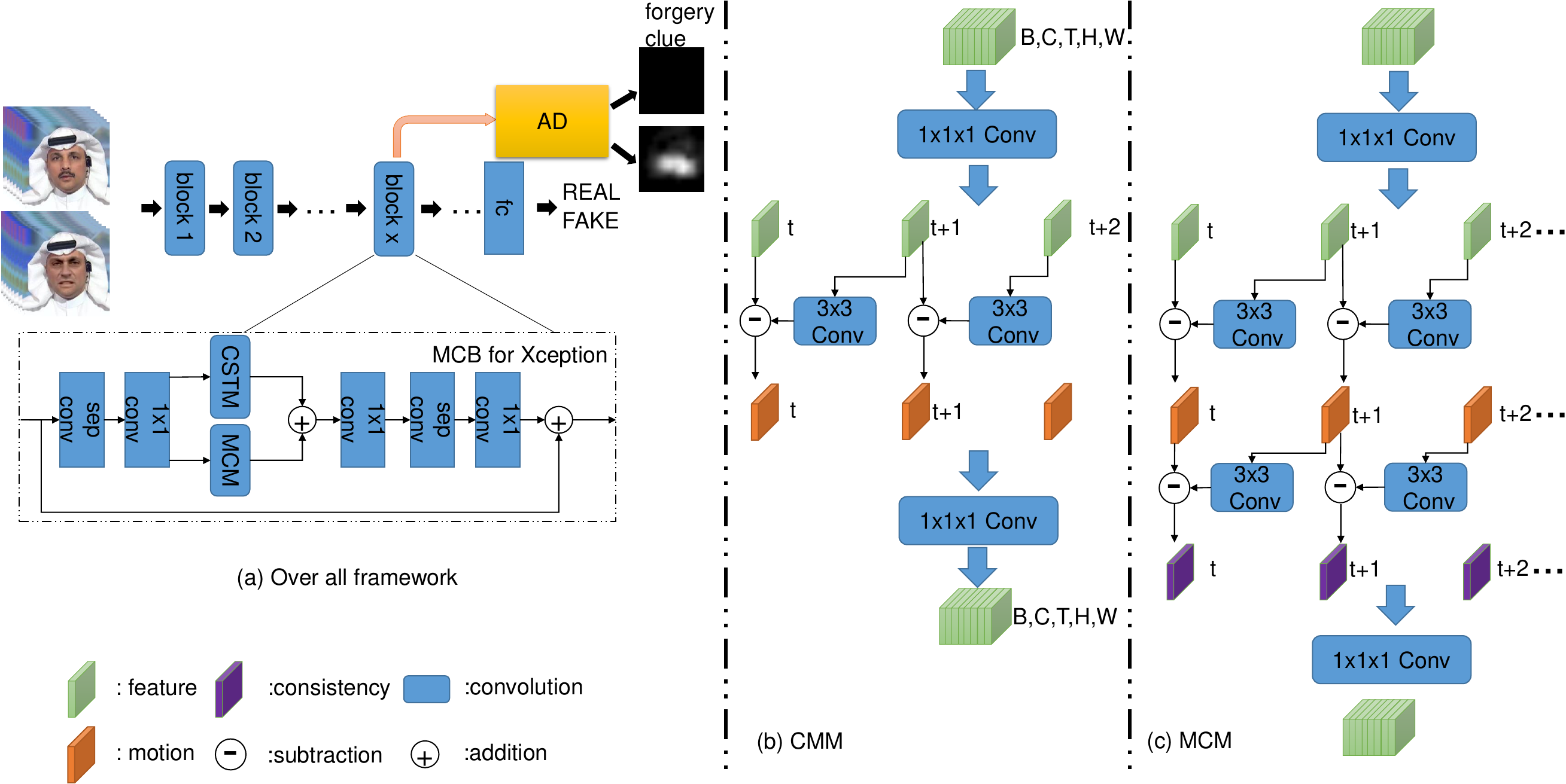} 
\caption{Architecture of the proposed Generalizable Motion Learning Network(a). The difference between Channel-wise Motion Module(b) and Motion Consistency Module(c).}
\label{fig2}
\end{figure*}

The manipulation clues left by different types of forgery methods in the motion field are diverse and unpredictable, while the real faces follow some fixed motion patterns.
Based on this observation, we propose an anomaly detection (AD) block and show abnormal motions are the essential cue to regularize the feature learning in the motion domain.
The whole framework of our proposed method is shown in Fig. \ref{fig2}(a).

\section{Methodology}
\label{sec:method}
\subsection{Preliminaries}

In this paper, we take the STM block proposed in \cite{STM} for its efficient spatio-temporal and motion encoding, while other blocks with explicit motion modeling can also be chosen. STM block consists of two modules, namely Channel-wise spatio-temporal Module(CSTM) and Channel-wise Motion Module(CMM).

By a temporal fusion operation and a spatial convolution operation, the CSTM extracts robust features from a sequence input.
Given a frame sequence feature map $\textbf{F} \in {\mathbb{R}} ^{B \times C \times T \times H \times W}$, the CSTM firstly performs frame fusion on adjacent frames by a channel-wise $3 \times 1 \times 1$ convolution.
After frame fusion, a $1 \times 3 \times 3$ channel-wise convolution is used for spatial embedding. We denote the spatio-temporal feature from CSTM as $\textbf{F}^{S}$.

Besides effective spatio-temporal modeling, STM proposes a Channel-wise Motion Module(CMM) for motion embedding.
As presented in Fig. \ref{fig2}(b), given a frame sequence feature map $\textbf{F} \in {\mathbb{R}} ^{B \times C \times T \times H \times W}$, the CMM firstly down-samples the channels of $\textbf{F}$ to $\textbf{F}^{D} \in {\mathbb{R}} ^{B \times C/16 \times T \times H \times W}$ by a $1 \times 1 \times 1$ convolution kernel for  computation reduction. Then, the CMM uses a $1 \times 3 \times 3$ channel-wise convolution $K^{PM}$ to model the motion of each pixel on $\textbf{F}^{D}$ and generates the pre-modeled motion feature $\textbf{F}^{PM}$.
Finally, the motion feature $\textbf{F}^{M}$ is generated through a frame by frame subtraction on $\textbf{F}^{D}$ and $\textbf{F}^{PM}$:
\begin{equation}
\begin{split}
\textbf{F}^{PM}_{b,c,t,h,w}=\sum_{i,j \in \{-1,0,1\}}{K^{PM}_{c,i,j}\textbf{F}^{D}_{b,c,t,h+i,w+j}}\\
\textbf{F}^{M}_{b,c,t,h,w}=\textbf{F}^{D}_{b,c,t,h,w} - \textbf{F}^{PM}_{b,c,t-1,h,w}
\end{split}
\label{eq1}
\end{equation}
Inspired by STM, our initial work transforms the spatial convolution kernels in a frame-based backbone into STM blocks and applies a binary classifier to face forgery detection.
A cross-entropy loss($L^{cls1}$) is used to constraint the classification results.

\subsection{Motion Consistency Block}
Even though STM block provides expressive spatio-temporal and motion modeling, it does not consider the motion consistency information since it aims to solve the action classification problem. However, the motion consistency information plays a key role in face forgery detection.
Therefore, we propose our Motion Consistency Block(MCB) which uses a Motion Consistency Module(MCM) to substitute the CMM in STM to capture the motion consistency information explicitly, while keeping the original CSTM in STM for general spatio-temporal features extracting.
MCM shown in Fig.\ref{fig2}(c) attempts to embed the motion relationship and motion consistency of pixels between adjacent frames by a second-order differential architecture.
\begin{equation}
\begin{split}
\textbf{F}^{PMM}_{b,c,t,h,w}=\sum_{i,j \in \{-1,0,1\}}{K^{PM}_{c,i,j}\textbf{F}^{M}_{b,c,t,h+i,w+j}}\\
\textbf{F}^{MM}_{b,c,t,h,w}=\textbf{F}^{M}_{b,c,t,h,w} - \textbf{F}^{PMM}_{b,c,t-1,h,w}
\end{split}
\label{eq2}
\end{equation}
As presented in Eq.\ref{eq2}, MCM uses $K^{PM}$ again to model the pixel motion $\textbf{F}^{M}$ and outputs the pre-modeled consistency feature $\textbf{F}^{PMM}$.
A frame by frame subtraction is performed on $\textbf{F}^{M}$ and $\textbf{F}^{PMM}$ to get $\textbf{F}^{MM}$ as the consistency of pixel motion.
We insist that $\textbf{F}^{M}$ and $\textbf{F}^{MM}$ are helpful to face forgery detection since the face manipulation methods usually work on independent frames and they do not take the coordination and consistency of motion face into consideration.
Finally, the output of MCB is the sum of $\textbf{F}^{S}$, $\textbf{F}^{M}$, and $\textbf{F}^{MM}$.

\subsection{Anomaly Detection Auxiliary Block}
Inspired by Face X-ray\cite{2020Face} which proposes blending boundary to regularize the feature learning in the spatial domain and makes the learned features generalize well, we propose abnormal motion is the essential cue
to regularize the feature learning in the motion domain.

Since the real faces follow some fixed motion patterns, while the motion patterns of fake faces are messy and unpredictable, we should force the feature distribution of the natural faces to be compact, and not constrain the features of the fake ones.
Specifically, we use the proposed AD block to get the motion maps which takes the transformed spatio-temporal features  $\textbf{F}^{out}$ from the backbone as input, as shown in Fig.\ref{fig2}(a). The AD block has a residual architecture and consists of nine rest convolution blocks. It produces candidate forgery clues $F^{*}$ which has the same size with $\textbf{F}^{out}$. We use the L1 loss($L^{l1}$) to make the forgery clues of real faces to be a zero-map:
\begin{equation}
L^{l1} =\frac{1}{N^P} \sum_{{F^{*}_i} \in  positive} ||F^{*}_i||_{1}
\end{equation}
where $N^P$ is the number of positive samples.
However, with only the L1 loss, there is a risk that AD outputs a zero-map for all samples, which makes this module learn nothing. To avoid this risk, forgery clues $F^{*}$ will go through a linear classifier layer, and an additional cross-entropy loss($L^{cls2}$) is adopted to make the forgery clues of fake faces apart from the real ones.
Our AD is regarded as an auxiliary branch.
We use the sum of the three losses as the final loss: $Loss=L^{cls1}+ L^{L1} + L^{cls2}$.
The classifiers of the backbone and the AD branch both give a forgery prediction, we use the average of them during the inference time.

\begin{table}
\setlength{\tabcolsep}{4pt}
\begin{center}
\caption{Evaluation ACC on the level-5 random distortion version DeeperForensics-V1.0. Our model is trained on C23 version of FF++(first row) and DFO without any data distortion(second row).
}
\label{table2}
\begin{tabular}{l|llllll|l}
\hline\noalign{\smallskip}
Source & Xce- & I3D &C3D  &R3D &R50+ &TSN &GML \\
Domain & pion\cite{Xception}  &\cite{2017Quo}  &\cite{2014Learning}  &\cite{2018A} &lstm &\cite{2016Temporal} &(Ours) \\
 \hline
FF++ &80.2 & 78.6  & - & 56.4 & - &-  &\textbf{85.8}\\
DFO &88.0 & 90.8  & 87.6 & 96.5 & 90.6 &91.5 &\textbf{99.9}\\
\noalign{\smallskip}
\hline
\noalign{\smallskip}
\hline
\end{tabular}
\end{center}
\vspace{-0.5cm}
\end{table}
\setlength{\tabcolsep}{1.4pt}
\section{Experiment}
\label{sec:experiment}

FaceForensics++\cite{FFPP} is a widely used face forgery detection benchmark, which contains 1000 original real videos from YouTube and each real video corresponds to 4 forgery methods.
We train our model on the HQ version of FF++ and evaluate it on two other popular forgery detection datasets\cite{DFO,CDF} to show the generalization ability of our model.
We adopt Area Under Receiver Operating Characteristic Curve(AUC) and Accuracy(ACC) as the evaluation metrics for extensive experiments.
We use a popular face extractor dlib\cite{2009Dlib} to detect faces.
Supposing $w$ and $h$ are the face width and height of the first frame respectively, we crop the region centered at the face in the initial frame with the size of $2\sqrt{wh} \times 2\sqrt{wh}$.
For each video, we save the cropped images with a size of $224 \times 224$ every 2 frames.
We collect every 8 frames as a sequence input.
Our models are trained with a SGD optimizer with a learning rate of 0.001 and weight decay of 1e-6.
We train our models with 150000 iterations and a bath size of 16 on a Nvidia Tesla-V100 GPU.
\subsection{Evaluation on DeeperForensics-V1.0}
DeeperForensics-V1.0\cite{DFO}(DFO) contains the same 1000 original videos as FF++, however, it uses a new face generation method based on auto-encoders for face swaps and provides several data distortion versions.
We conduct two types of cross-domain evaluation on the DFO dataset.
The first type is cross dataset evaluation(first row of Table \ref{table2}).
We train our model on the HQ version of FF++ and evaluate it on the test set of DFO.
With the help of MCB and AD block, our method gains an ACC increase of over $5.6\%$ based on the Xception baseline.
Compared with the famous sequence-based methods(I3D\cite{2017Quo}, C3D\cite{2014Learning}, and R3D\cite{2018A}), our model shows the best generalization ability with unseen forgery methods and data distortions.
The second type is cross-data distortion evaluation(second row of Table \ref{table2}).
We follow the std $\to$ std/sing strategy recommended in \cite{DFO} which is more like a cross-distortion testing.
The results in Table \ref{table2}(second row) demonstrate the excellent generalization performance of the proposed method on cross-data distortion evaluation of DFO.
It achieves the best performance compared with the most famous sequence-based backbones and outperforms its baseline Xception by $11.9\%$.

\subsection{Evaluation on Celeb-DF}
Celeb-DF\cite{CDF} contains 540 real videos collected from YouTube and 5639 fake videos which are more difficult to distinguish than the fake faces in the previous datasets.
In intra-domain evaluation(first row of Table \ref{table3}),
our model achieves the best performance in terms of AUC score.
Some other state-of-art methods(Xception\cite{Xception}, F$^{3}$-Net\cite{qian2020thinking}, Effb4\cite{effnet}, Matt\cite{Matt}, Recc\cite{RECCE}) also show competitive performance. However, our model shows outstanding generalizability compared with them.
We train our model on the HQ version of FF++\cite{FFPP} and evaluate it on the test set of CDF.
The second row of Table \ref{table3} demonstrates the generalizability of the proposed method.
It outperforms the baseline by about $10\%$ in cross-domain evaluation. Our method achieves the best performance both in intra-domain and cross-domain evaluation on CDF.

\setlength{\tabcolsep}{4pt}
\begin{table}
\begin{center}
\caption{Evaluation AUC on Celeb-DF. The models are trained on CDF(first row) and C23 version of FF++(second row),respectively.}
\label{table3}
\begin{tabular}{l|llllll|l}
\hline\noalign{\smallskip}
Source & Xce-  &Recc &Tbra &F$^3$- &Effb4 &Matt &GML\\
Domain & pion &\cite{RECCE} &\cite{2020Two} &Net\cite{qian2020thinking} &\cite{effnet} &\cite{Matt} &(Ours)\\
 \hline
CDF &99.7  & 99.9 &93.2 & 98.1 &99.7  & 99.8 &\textbf{99.9}\\
FF++ &65.5 & 68.7 &73.4 & 65.1 &64.3 &67.4 &\textbf{75.3}\\
\noalign{\smallskip}
\hline
\noalign{\smallskip}
\hline

\end{tabular}
\end{center}
\vspace{-0.4cm}
\end{table}
\setlength{\tabcolsep}{1.4pt}
\subsection{Ablation Study}
\setlength{\tabcolsep}{4pt}
\begin{table}
\begin{center}
\caption{Cross-dataset evaluation, model trained from FF++}
\label{table4}
\begin{tabular}{l|ll}
\hline\noalign{\smallskip}
Target Domain &DFO &CDF \\
\hline
Xception &60.0 &65.5\\
Xception+CSTM &62.5 &65.3\\
Xception+STM &73.0 &70.0 \\
Xception+MCB &77.0 &72.0 \\
Xception+AD &60.0 &64.0 \\
Xception+STM+AD &79.3 &69.2 \\
Xception+MCB+AD$_{wocls}$ &77.5 &71.6 \\
Xception+MCB+AD$_{wol1}$ &78.3 &72.3 \\
\hline
Xception+MCB+AD &\textbf{85.8} &\textbf{75.3} \\

\noalign{\smallskip}
\hline
\noalign{\smallskip}
\hline
\end{tabular}
\end{center}
\vspace{-0.4cm}
\end{table}
\setlength{\tabcolsep}{1.4pt}

\begin{figure}[t]
\centering
\vspace{-0.4cm}
\includegraphics[width=0.9\columnwidth]{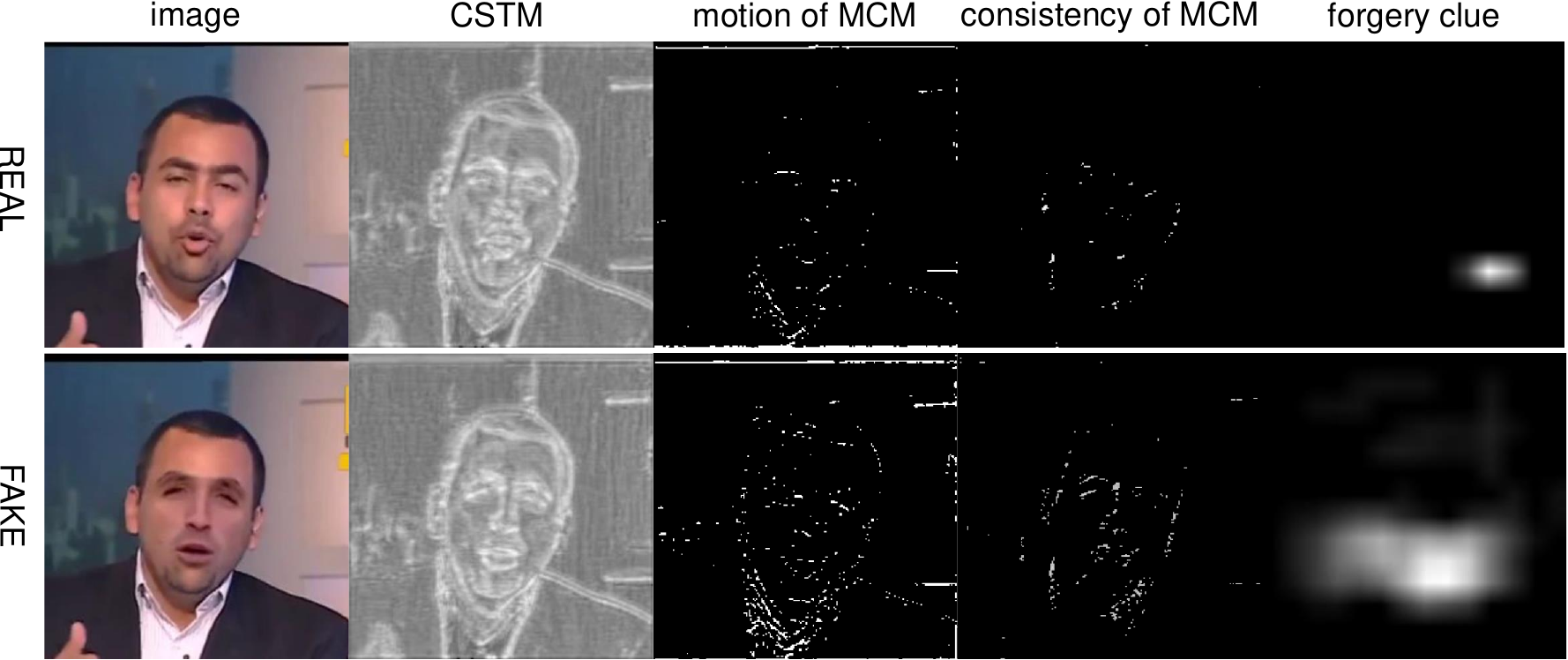}
\caption{Visualization of heatmaps from our model.
}
\label{fig4}
\vspace{-0.4cm}
\end{figure}

Xception+STM(consists of CSTM and CMM) and Xception+MCB
(consists of CSTM and MCM) in Table \ref{table4} mean that we replace the depth-wise convolution of Xception backbone with STM block and our MCB as stated in the methodology section respectively, and the AD is not added to the framework. The performance in Table \ref{table4} shows that on the FF++ to DFO test, our MCB boosts the Xception backbone with an AUC improvement of $17\%$.
While on the FF++ to CDF cross-set test, our MCB improves the AUC of the Xception backbone by $6.5\%$.
Compared with the basic STM block, our MCB block gains improvement of $4\%$ and $2\%$ on DFO and CDF, respectively, which shows the effectiveness of the motion consistency modeling.
As presented in Fig. \ref{fig4}, the CSTM extracts textual information.
Our MCM provides motion information(col3) and the fake faces contain more motion information.
Besides, our MCM extracts additional motion consistency information(col4) and the fake faces show more motion inconsistency.

To show the effectiveness of our AD module for generalizable motion feature learning, we add AD model into the 5th block of the Xception, the Xception+STM and Xception+MCB frameworks, separately.
As presented in  Table \ref{table4},
our AD shows no improvement on frame-based Xception backbone. However, our AD works on sequence-based networks i.e. the STM and MCB framework.
Our AD improves the AUC of Xception+MCB framework by $8.8\%$ and $3.3\%$ on the FF++ to DFO and CDF cross-domain test respectively. This demonstrates that abnormal motion is an effective cue to regularize the feature learning.
Visualization of the fifth column Fig.\ref{fig4} proves that fake faces leave more clues in AD heatmaps.

To prove the effectiveness of the losses in the AD module, we removed the CrossEntropy loss(Xception+MCB+AD$_{wocls}$) and L1 loss(Xception+MCB+AD$_{wol1}$), separately.
If we remove the L1 loss, the performance drops, since there are no constraints on the learned feature maps. This also verifies that the improvement comes from the AD module design which forces it to learn motion anomaly-related features rather than the ensemble model.
The performance also drops and is close to the one without the AD module if we remove the CrossEntropy loss.
The CrossEntropy loss avoids this risk of outputting a zero-amp for all samples.

\section{Conclusion}
\label{sec:Conclusion}

We propose an sequence-based forgery detection framework based on an existing video classification block.
A motion consistency block is proposed to perform motion consistency embedding for effective forgery detection.
An anomaly detection block with the combining loss makes the spatio-temporal features more generalizable.
In future work, we will apply the proposed framework to other video vision tasks.

\bibliographystyle{IEEEbib}
\bibliography{refs}

\begin{thebibliography}{10}

\bibitem{FFPP}
Andreas Rossler, Davide Cozzolino, Luisa Verdoliva, Christian Riess, Justus
  Thies, and Matthias Nie{\ss}ner,
\newblock ``Faceforensics++: Learning to detect manipulated facial images,''
\newblock in {\em ICCV}, 2019, pp. 1--11.

\bibitem{DFO}
Liming Jiang, Ren Li, Wayne Wu, Chen Qian, and Chen~Change Loy,
\newblock ``Deeperforensics-1.0: A large-scale dataset for real-world face
  forgery detection,''
\newblock in {\em CVPR}, 2020, pp. 2889--2898.

\bibitem{CDF}
Yuezun Li, Xin Yang, Pu~Sun, Honggang Qi, and Siwei Lyu,
\newblock ``Celeb-df: A large-scale challenging dataset for deepfake
  forensics,''
\newblock in {\em CVPR}, 2020, pp. 3207--3216.

\bibitem{sun2021domain}
Ke~Sun, Hong Liu, Qixiang Ye, Jianzhuang Liu, Yue Gao, Ling Shao, and Rongrong
  Ji,
\newblock ``Domain general face forgery detection by learning to weight,''
\newblock in {\em AAAI}, 2021, vol.~35, pp. 2638--2646.

\bibitem{Xception}
Fran{\c{c}}ois Chollet,
\newblock ``Xception: Deep learning with depthwise separable convolutions,''
\newblock in {\em CVPR}, 2017, pp. 1251--1258.

\bibitem{Matt}
Hanqing Zhao, Wenbo Zhou, Dongdong Chen, Tianyi Wei, Weiming Zhang, and Nenghai
  Yu,
\newblock ``Multi-attentional deepfake detection,''
\newblock in {\em CVPR}, June 2021, pp. 2185--2194.

\bibitem{qian2020thinking}
Yuyang Qian, Guojun Yin, Lu~Sheng, Zixuan Chen, and Jing Shao,
\newblock ``Thinking in frequency: Face forgery detection by mining
  frequency-aware clues,''
\newblock in {\em ECCV}. Springer, 2020, pp. 86--103.

\bibitem{chen2021local}
Shen Chen, Taiping Yao, Yang Chen, Shouhong Ding, Jilin Li, and Rongrong Ji,
\newblock ``Local relation learning for face forgery detection,''
\newblock in {\em Proceedings of the AAAI Conference on Artificial
  Intelligence}, 2021, vol.~35, pp. 1081--1088.

\bibitem{zhang2019detecting}
Xu~Zhang, Svebor Karaman, and Shih-Fu Chang,
\newblock ``Detecting and simulating artifacts in gan fake images,''
\newblock in {\em 2019 IEEE International Workshop on Information Forensics and
  Security (WIFS)}. IEEE, 2019, pp. 1--6.

\bibitem{2018Deepfake}
David G{\"u}era and Edward~J Delp,
\newblock ``Deepfake video detection using recurrent neural networks,''
\newblock in {\em proceedings of the IEEE international conference on advanced
  video and signal based surveillance}. IEEE, 2018, pp. 1--6.

\bibitem{2020Deepfake}
Oscar de~Lima, Sean Franklin, Shreshtha Basu, Blake Karwoski, and Annet George,
\newblock ``Deepfake detection using spatiotemporal convolutional networks,''
\newblock {\em CoRR}, vol. abs/2006.14749, 2020.

\bibitem{2020Spatio}
Ipek Ganiyusufoglu, L.~Minh Ng{\^{o}}, Nedko Savov, Sezer Karaoglu, and Theo
  Gevers,
\newblock ``Spatio-temporal features for generalized detection of deepfake
  videos,''
\newblock {\em CoRR}, vol. abs/2010.11844, 2020.

\bibitem{WildDeepfake}
Bojia Zi, Minghao Chang, Jingjing Chen, Xingjun Ma, and Yu{-}Gang Jiang,
\newblock ``Wilddeepfake: {A} challenging real-world dataset for deepfake
  detection,''
\newblock in {\em {MM} '20: The 28th {ACM} International Conference on
  Multimedia, 2020}, 2020, pp. 2382--2390.

\bibitem{STM}
Boyuan Jiang, MengMeng Wang, Weihao Gan, Wei Wu, and Junjie Yan,
\newblock ``Stm: Spatiotemporal and motion encoding for action recognition,''
\newblock in {\em ICCV}, 2019, pp. 2000--2009.

\bibitem{2020Face}
Lingzhi Li, Jianmin Bao, Ting Zhang, Hao Yang, Dong Chen, Fang Wen, and Baining
  Guo,
\newblock ``Face x-ray for more general face forgery detection,''
\newblock in {\em CVPR}, 2020, pp. 5001--5010.

\bibitem{2017Quo}
Joao Carreira and Andrew Zisserman,
\newblock ``Quo vadis, action recognition? a new model and the kinetics
  dataset,''
\newblock in {\em CVPR}, 2017, pp. 6299--6308.

\bibitem{2014Learning}
Du~Tran, Lubomir Bourdev, Rob Fergus, Lorenzo Torresani, and Manohar Paluri,
\newblock ``Learning spatiotemporal features with 3d convolutional networks,''
\newblock in {\em ICCV}, 2015, pp. 4489--4497.

\bibitem{2018A}
Du~Tran, Heng Wang, Lorenzo Torresani, Jamie Ray, Yann LeCun, and Manohar
  Paluri,
\newblock ``A closer look at spatiotemporal convolutions for action
  recognition,''
\newblock in {\em CVPR}, 2018, pp. 6450--6459.

\bibitem{2016Temporal}
Limin Wang, Yuanjun Xiong, Zhe Wang, Yu~Qiao, Dahua Lin, Xiaoou Tang, and Luc
  Van~Gool,
\newblock ``Temporal segment networks: Towards good practices for deep action
  recognition,''
\newblock in {\em ECCV}. Springer, 2016, pp. 20--36.

\bibitem{2009Dlib}
Davis~E King,
\newblock ``Dlib-ml: A machine learning toolkit,''
\newblock {\em The Journal of Machine Learning Research}, vol. 10, pp.
  1755--1758, 2009.

\bibitem{effnet}
Mingxing Tan and Quoc Le,
\newblock ``Efficientnet: Rethinking model scaling for convolutional neural
  networks,''
\newblock in {\em International Conference on Machine Learning}. PMLR, 2019,
  pp. 6105--6114.

\bibitem{RECCE}
Junyi Cao, Chao Ma, Taiping Yao, Shen Chen, Shouhong Ding, and Xiaokang Yang,
\newblock ``End-to-end reconstruction-classification learning for face forgery
  detection,''
\newblock in {\em CVPR}, 2022, pp. 4113--4122.

\bibitem{2020Two}
Iacopo Masi, Aditya Killekar, Royston~Marian Mascarenhas, Shenoy~Pratik
  Gurudatt, and Wael AbdAlmageed,
\newblock ``Two-branch recurrent network for isolating deepfakes in videos,''
\newblock in {\em ECCV}. Springer, 2020, pp. 667--684.

\end{thebibliography}

\end{document}